\documentclass[runningheads]{llncs}
\usepackage{graphicx}
\usepackage{graphics}
\usepackage{graphicx}
\usepackage{tikz}

\usepackage{url}
\usepackage{xcolor}
\RequirePackage[colorlinks,citecolor=blue,urlcolor=blue]{hyperref}

\begin{document}

\def\checkmark{\tikz\fill[scale=0.4](0,.35) -- (.25,0) -- (1,.7) --
(.25,.15) -- cycle;} 
\def\todoref{\footnote{TODO ref}}
\def\furl#1{\footnote{\url{#1}}}
\def\Tref#1{Table~\ref{#1}}
\def\Fref#1{Figure~\ref{#1}}
\def\hideXXX#1{}
\def\fXXX#1{\footnote{\XXX{#1}}}
\def\XXX#1{}

\def\citet#1{\newcite{#1}}
\def\citep#1{\cite{#1}}
\def\parcite#1{\cite{#1}}
\def\inparcite#1{\newcite{#1}}
\def\perscite#1{\newcite{#1}}

\def\samp#1{``\textit{#1}''}

\def\plusminus{$\pm$}

\def\stefull{Sub\discretionary{-}{}{}word\discretionary{-}{}{}Text\discretionary{-}{}{}En\discretionary{-}{}{}coder} 
\def\ste{STE} 
\def\gsw{\ste}
\def\gswfull{\stefull}
\def\ttot{Tensor2Tensor}

\hyphenation{Deri-Net}

\def\orcidID#1{\href{https://orcid.org/#1}{\includegraphics[height=1em]{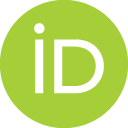}}}

%
\title{Morphological and Language-Agnostic \\Word Segmentation for
NMT\thanks{
This work has been supported by
the grants
18-24210S of the Czech Science Foundation, 
SVV~260~453
and ``Progress'' Q18+Q48 of Charles University,
H2020-ICT-2014-1-645452 (QT21) of the EU, 
and using language resources distributed by
the LINDAT/CLARIN project of the Ministry
of Education, Youth and Sports of the Czech
Republic (projects LM2015071 and OP VVV
VI CZ.02.1.01/0.0/0.0/16 013/0001781).
We thank Jaroslava Hlav\'{a}\v{c}ov\'{a} for digitizing excerpts of \parcite{slavickova1975} used as gold-standard data for evaluating the segmentation methods.
}}

%
%

\author{Dominik Mach{\' a}{\v c}ek\orcidID{0000-0002-5530-1615} \and
Jon{\' a}{\v s} Vidra\orcidID{0000-0003-1470-9005} \and
Ond{\v r}ej Bojar\orcidID{0000-0002-0606-0050}}


%

\institute{ 
Charles University, Faculty of Mathematics and Physics, \\
Institute of Formal and Applied Linguistics, \\
Malostransk{\' e} n{\' a}m{\v e}st{\' i} 25, \\
118 00 Prague, Czech Republic \\
\email{\{machacek,vidra,bojar\}@ufal.mff.cuni.cz} \\
\url{http://ufal.mff.cuni.cz}
}

%
\maketitle              
\begin{abstract}
The state of the art of handling rich morphology in neural machine translation
(NMT) is to break word forms into subword units, so that the overall vocabulary
size of these units fits the practical limits given by the NMT model and GPU
memory capacity.
In this paper, we compare two common but linguistically uninformed methods of
subword construction (BPE and \ste, the method implemented in \ttot{} toolkit)
and two linguistically-motivated methods: Morfessor and one
novel method, based on a derivational dictionary.
Our experiments with German-to-Czech translation, both morphologically rich,
document that so far, the non-motivated methods perform better.
Furthermore, we identify a critical difference between BPE and \ste{} and show
a simple pre-processing step for BPE that considerably increases 
translation quality as evaluated by automatic measures.
\end{abstract}

\section{Introduction}

One of the key steps that allowed to apply neural machine translation (NMT) in
unrestricted setting was the move to subword units. While the natural (target) vocabulary size
in a realistic parallel corpus exceeds the limits imposed by model size
and GPU RAM, the vocabulary size of custom subwords can be kept small.

The current most common technique of subword construction is called byte-pair
encoding (BPE) by Sennrich et al.
\cite{BPE}.\furl{http://github.com/rsennrich/subword-nmt/} Its counterpart
originating in the
commercial field is wordpieces \parcite{google:bridging:gap:2016:arxiv}. Yet
another variant of the technique is implemented in Google's open-sourced toolkit
\ttot{},\furl{http://github.com/tensorflow/tensor2tensor} namely the \stefull{}
class (abbreviated as \ste{} below).

The common property of these approaches is that they are trained in an
unsupervised fashion, relying on the distribution of character sequences, but
disregarding any morphological properties of the languages in question.


On the positive side, BPE and \ste{} (when trained jointly for both the
source and target languages) allow to identify and benefit from words that share
the spelling in some of their part, e.g. the root of the English
\samp{legalization} and Czech \samp{legalizace} (noun) or
\samp{legaliza\v{c}n\'{i}} (adj).
On the downside, the 
root of different word forms of one lemma can be split in several different
ways
and the neural network will not explicitly know about their relatedness.
A morphologically motivated segmentation method could solve this issue by
splitting words into their constituent semantics- and syntax-bearing parts.


In this paper, we experiment with two methods aimed at morphologically adequate splitting of
words in a setting involving two morphologically rich languages: Czech and German.
We also compare the performance of several variations of BPE and \ste{}.
Performance is analysed both by intrinsic evaluation of morphological adequateness,
and extrinsically by evaluating the systems on a German-to-Czech translation task.


%
%
\section{Morphological Segmentation}



Huck et al. \cite{Huck}
benefit from linguistically aware separation of suffixes prior to BPE
on the target side of
medium-size English to German translation task (overall improvement about 0.8 BLEU).
Pinnis et al. \cite{Pinnis} show similar improvements with
analogical prefix and suffix splitting on English to Latvian.

Since there are no publicly available morphological segmentation tools for Czech,
we experimented with an unsupervised morpheme induction tool,
Morfessor 2.0 \citep{morfessor},
and we developed a simple supervised method based on derivational morphology.

\subsection{Morfessor}

Morfessor \citep{morfessor} is an unsupervised segmentation tool that
utilizes a probabilistic model of word formation.  The segmentation obtained
often resembles a linguistic morpheme segmentation, especially in
compounding languages, where Morfessor benefits from the uniqueness
of the textual representation of morphs. It can be used to split compounds,
but it is not designed to handle phonological and orthographical changes as
in Czech words \samp{\v{z}e\v{n}}, \samp{\v{z}n\v{e}} (\samp{harvest} in singular and
plural). In Czech orthography, adding plural suffix \samp{e} after
\samp{\v{n}} results in \samp{n\v{e}}. This suffix also causes phonological
change in this word, the first \samp{e} is dropped. 
Thus, \samp{\v{z}e\v{n}} and \samp{\v{z}n} are two variants of the same
morpheme, but Morfessor can't handle them appropriately.

\subsection{DeriNet}

Our novel segmentation method works by exploiting word-to-word relations
extracted from DeriNet \citep{derinet}, a network of Czech lexical derivations,
and MorfFlex \citep{morfflex}, a Czech inflectional dictionary.
DeriNet is a collection of directed trees of derivationally connected lemmas.
MorfFlex is a list of lemmas with word forms and morphological tags.
We unify the two resources by taking the trees from DeriNet as the basis and
adding all word forms from MorfFlex as new nodes (leaves) connected with
their lemmas.

The segmentation algorithm works in two steps: Stemming of words based on their
neighbours and morph boundary
propagation.


\def\edge#1#2{\samp{#1}$\rightarrow$\samp{#2}}
\def\edgetrans#1#2#3#4{\samp{#1} (#2)$\rightarrow$\samp{#3} (#4)}

We approximate stemming by detecting the longest common substring of each pair
of connected words. This segments both words connected by an edge into
a (potentially empty) prefix, the common substring and a (potentially empty)
suffix, using exactly two splits.
For example, the edge \edgetrans{m\'{a}vat}{to be waving}{m\'{a}vnout}{to wave}
has the longest common substring of \samp{m\'{a}v}, introducing the splits
\samp{m\'{a}v-at} and \samp{m\'{a}v-nout} into the two connected words.

Each word may get multiple such segmentations, because it may have more than
one word connected to it by an edge. Therefore, the stemming phase itself
can segment the word into its constituent morphs; but in the usual case,
a multi-morph stem is left unsegmented.
For example, the edge \edgetrans{m\'{a}vat}{to be waving}{m\'{a}vaj\'{i}c\'{i}}{waving}
has the longest common substring of \samp{m\'{a}va}, introducing the splits \samp{m\'{a}va-t}
and \samp{m\'{a}va-j\'{i}c\'{i}}. The segmentation of \samp{m\'{a}vat} is therefore
\samp{m\'{a}v-a-t}, the union of its splits based on all linked words.


To further split the stem, we propagate morph boundaries from connected words.
If one word of a connected pair contains a split in their common substring
the other word does not, the split is copied over. This way, boundaries are
propagated through the entire tree.
For example, we can split \samp{m\'{a}va-j\'{i}c\'{i}} further using the other
split in \samp{m\'{a}v-a-t} thanks to it lying in the longest common substring
\samp{m\'{a}va}. The segmentation of \samp{m\'{a}vaj\'{i}c\'{i}} is therefore
\samp{m\'{a}v-a-j\'{i}c\'{i}}.

These examples also shows the limitations of this method: the words are often
split too eagerly, resulting in many single-character splits. The boundaries
between morphemes are fuzzy in Czech because connecting phonemes
are often inserted and phonological changes occur. These cause spurious or
misplaced splits. For example, the single-letter morph \emph{a} in
\emph{m\'{a}v-a-t} and \emph{m\'{a}v-a-j\'{i}c\'{i}} does not carry any information useful in
machine translation and it would be better if we could detect it as a
phonological detail and leave it connected to one of the neighboring morphs.

\section{Data-Driven Segmentation}

%
%
%



We experimentally compare BPE with STE.
As we can see in the left side of
\Fref{fig:splits}, a distinct feature of \ste{} seems to be an
underscore as a zero
suffix mark appended to every word before the subword splits are determined.
This small trick allows to learn more adequate units compared to BPE.
For example, the Czech word form
\samp{tramvaj} (\samp{a tram}) can serve as a subword unit that, combined with
zero suffix (\samp{\_}) corresponds to the
nominative case or, combined with the suffix
\samp{e} to the genitive case \samp{tramvaje}.
In BPE, there can be either \samp{tramvaj} as a standalone word or
two subwords \samp{tramvaj@@} and \samp{e} (or possibly split further) with no
vocabulary entry sharing
possible.

\def\star{(*)}
\def\placestar{\star}
\begin{figure}[t]
\centering
\scalebox{0.7}{
\begin{tabular}{ll|ll}
& \textbf{Language agnostic}
& & \textbf{Linguistically motivated} \\
\hline
\textbf{Tokenized} &
Bl{\' i}{\v z}{\' i} se k tob{\v e} tramvaj . 
& \textbf{DeriNet} & 
Bl@@ {\' i}{\v z}@@ {\' i} se k tob{\v e} tramvaj . \\
\placestar{} & Z tramvaje nevystoupili . &
\placestar{} & Z tram@@ vaj@@ e nevyst@@ oup@@ ili .
\\ \hline

\textbf{\ste{}} &
Bl{\'i}{\v z}{\'i}\_ se\_ k\_ tob\v{e}\_ tramvaj \_ .\_ & 
\textbf{DeriNet} & 
Bl@@ {\' i}{\v z}@@ {\' i}\_ se\_ k\_ tob{\v e}\_ tramvaj\_ .\_  \\

& Z\_ tramvaj e\_ nevysto upil i\_ .\_ &
\textbf{+\gsw{}} & Z\_ tra m@@ vaj@@ e\_ nevyst@@ oup@@ ili\_ .\_ 
\\
\hline

\textbf{BPE} &
Bl{\'i}{\v z}{\'i} se k tob{\v e} tram@@ vaj . 
& \textbf{Morfessor} &
Bl{\' i}{\v z}{\' i} se k tob{\v e} tramvaj .
\\
& Z tram@@ va@@ je nevy@@ stoupili . 
&
\placestar{} & Z tramvaj@@ e ne@@ vystoupil@@ i .
\\ \hline

\textbf{BPE und} &
Bl{\'i}{\v z}{\'i}\_ se\_ k\_ tob\v{e}\_ tram@@ vaj\_ .\_ &
\textbf{Morfessor} &
Bl{\' i}{\v z}{\' i}\_ se\_ k\_ tob{\v e}\_ tramvaj\_ .\_ 
\\ 
&
Z\_ tram@@ va@@ je\_ nevy@@ stoupili\_ .\_ 
&
\textbf{+\gsw{}}&
Z\_ tramvaj@@ e\_ ne@@ vystoupil@@ i\_ .\_
\\ 

\hline

\textbf{BPE und} &
Bl{\'i}{\v z}{\'i}\_ se\_ k\_ tob\v{e}\_ tram@@ vaj\_ . \\
\textbf{non-final}& Z\_ tram@@ va@@ je\_ nevy@@ stoupili\_ . \\ 

\end{tabular}
}
\caption{Example of different kinds of segmentation of Czech sentences \samp{You're
being approached by a tram. They didn't get out of a tram.}
Segmentations marked with \star{} are preliminary, they cannot be used in MT
directly alone because they do not
restrict the total number of subwords to the vocabulary size limit.
}
\label{fig:splits}
\end{figure}

To measure the benefit of this zero suffix feature, we modified BPE by appending an underscore prior to BPE training
in two flavours: (1) to every word (``BPE und''), and (2) to every word
except of the last word in the sentence (``BPE und non-final''). 


Another typical feature of \ste{} is to share the vocabulary of the source and target sides.
While there are almost no common words in Czech and German apart from digits, punctuation
and some proper names, it turns out that around 30\% of the
\ste{} shared German-Czech vocabulary still appears in both languages. This
contrasts to only 7\% of accidental 
overlap of separate BPE vocabularies.

\section{Morphological Evaluation}

\subsection{Supervised Morphological Splits}

We evaluate the segmentation quality in two ways: by looking at the data
and finding typical errors and by comparing the outputs of individual systems
with gold standard data from a printed dictionary of Czech morpheme segmentations \parcite{slavickova1975}.
We work with a sample of the book \parcite{slavickova1975} containing 14\,581 segmented verbs transliterated into modern Czech,
measuring precision and recall on morphs and morph boundaries and accuracy of totally-correctly segmented words.

\subsection{Results}

\Fref{fig:splits} shows example output on two Czech sentences.
The biggest difference between our DeriNet-based approach and Morfessor is
that Morfessor does not segment most stems at all, but in contrast to our system,
it reliably segments inflectional endings and the most common affixes.
The quality of our system depends on the quality of the underlying data.
Unfortunately, trees in DeriNet are not always complete, some derivational links are missing.
If a word belongs to such an incomplete tree, our system
will not propose many splits.
None of the methods handles phonological and orthographical changes,
which also severely limits their performance on Czech.

The results against golden Czech morpheme segmentations are in Table~\ref{tab:segment-quality}.

The scores on boundary detection seem roughly comparable, with different systems
making slightly different tradeoffs between precision and recall.
Especially the DeriNet-enhanced STE (``DeriNet+STE'') system sacrifices some precision
for higher recall.
The evaluation of morph detection varies more, with the best system being the standard BPE,
followed by BPE with shared German and Czech vocab.
This suggests that adding the German side to BPE decreases segmentation quality of Czech from the morphological point of view. 

The scores on boundary detection are necessarily higher than on morph detection, because
a correctly identified morph requires two correctly identified boundaries
--- one on each side.

Overall, the scores show that none of the methods
presented here is linguistically adequate. Even
the best setup reaches only 62\% F1 in boundary detection which translates to meagre 0.77\%
of all words in our test set without a flaw.

\begin{table}[t]
\caption{Morph segmentation quality on Czech as measured on gold standard data.}
\label{tab:segment-quality}
\centering
{
\scriptsize
\begin{tabular}{l|rrr|rrr|r}
& \multicolumn{3}{c|}{\textbf{Morph Detection}} & \multicolumn{3}{c|}{\textbf{Boundary Detection}} & \textbf{Word} \\
\textbf{Segmentation} & \textbf{Precision} & \textbf{Recall} & \textbf{F1} & \textbf{Precision} & \textbf{Recall} & \textbf{F1} & \textbf{Accuracy} \\ \hline
BPE&21.24&12.74&15.93&77.38&52.44&62.52&0.77 \\
BPE shared vocab&19.99&11.75&14.80&77.04&51.49&61.72&0.69 \\
\ste&13.03&7.79&9.75&77.08&51.77&61.93&0.23 \\
\ste+Morfessor&11.71&7.59&9.21&74.49&52.85&61.83&0.23 \\
\ste+DeriNet&13.89&10.44&11.92&70.76&55.00&61.89&0.35 \\
\end{tabular}
}

\end{table}

\section{Evaluation in Machine Translation}

\subsection{Data}

Our training data consist of Europarl v7
\citep{Europarl}
and OpenSubtitles2016 \citep{OPUS}, after some further cleanup.
Our final training corpus, processed with the Moses tokenizer
\parcite{moses}, consists of 8.8M parallel sentences, 89M tokens on the source
side, 78M on the target side. The vocabulary size is 807k and 953k on the source and
target, respectively.



We use WMT\furl{http://www.statmt.org/wmt13} newstest2011 as the development set and 
newstest2013 as the test set, 3k sentence pairs each.

All experiments were carried out in \ttot{}
(abbreviated as T2T), version
1.2.9,\furl{http://github.com/tensorflow/tensor2tensor} using the model
\texttt{transformer\_big\_single\_gpu}, batch size of 1500 and
\texttt{learning\_rate\_warmup\_steps} set to 30k or 60k if the learning
diverged.

The desired vocabulary size of subword units is set to 100k when shared for both source and target and to 50k
each with separate vocabularies.

Since T2T \stefull{} constructs the subword model only from a sample of the
training data, we had to manually set the
\texttt{file\_byte\_budget} variable in the code to 100M, otherwise not enough
distinct wordforms were observed to fill up the intended 100k vocabulary size.

For data preprocessed by BPE, we used T2T TokenTextEncoder which allows to use a
user-supplied vocabulary.

Final scores (BLEU,
CharacTER, chrF3 and BEER) are measured after removing any subword splits
and detokenizing with Moses detokenizer. Each of the metric implementation
handles tokenization on its own.

Machine translation for German-to-Czech language pair is currently
underexplored. We included Google Translate (as of May 2018, neural) into our evaluation and conclude the latest Transformer model has easily outperformed it on the given test dataset.

Due to a limited number of GPU cards, we
cannot afford multiple training runs for estimating statistical significance.
We at least report the average score of the test set as translated by several
model checkpoints around the same number of training steps where the BLEU score has already flattened. This
happens to be approximately after 40 hours of training around 300k training steps. 

\subsection{Experiment 1: Motivated vs. Agnostic Splits}

\begin{table}[t]
\caption{Data characteristics and
automatic metrics after 300k steps of training. 
}
\label{tab:dervoc}
\def\multic#1#2#3{\multicolumn{#1}{#2}{#3}}
\def\cros{\rlap{$^*$}}
\def\ul{}
%
\centering
\hideXXX{zakomentovana cesta k datum:}

\hideXXX{DM: mam jeste dva behy, kde byl STE natrenovan na dvakrat vic datech
budto ze zdroje, nebo z cile. Trenovaci krivka BLEU je shodna s baseline.
Zvazim pridani a to, jestli tam neni chyba.}
\hideXXX{DM: update: experimenty se nepovedly, segmentace je stejna jako
u baseline, protoze jsem spatne vytvoril dataset na natrenovani STE. On
bere jenom zacatek, tech 100M tokenu. Nekdy priste...
Sem ale muzu pozdeji pridat BEER, ChrF3, nCharacTER, vejde se.
}

\scriptsize
\def\ccol#1{\multicolumn{1}{c}{\textbf{#1}}}%
\def\ccoll#1{\multicolumn{1}{c|}{\textbf{#1}}}%
\begin{tabular}{ll|rr|rrr|ccccccccc}
\multicolumn{2}{l|}{}  &  \multicolumn{2}{c|}{\textbf{tokens}} &  \multicolumn{2}{c}{\textbf{types}}  &  \textbf{\%}    \\
\ccol{de}           &  \multicolumn{1}{c}{\textbf{cs}}
& \multicolumn{1}{|c}{\textbf{de}} &  \ccoll{cs}  &   \ccol{de}  &  \ccol{cs}  &  \textbf{shrd}  &  \ccol{BLEU} &  \ccol{CharacTER}  &  \ccol{chrF3}  &  \ccol{BEER}  \\  \hline
\ste{}       &  \ste{}       &  97M   &  87M   &  54k  &  74k  &  29.89  &  18.78 & 61.27 & 47.82& 50.34 \\
\ste{}       &  Morfessor+\ste{}  &  95M   &  98M   &  63k  &  63k  &  26.42  &  18.22 & 62.27 & 47.30& 50.00 \\
Morfessor+\ste{}  &  DeriNet+\ste{}   &  138M  &  308M  &  63k  &  69k  &  36.82  &  16.99 & 64.26 & 45.64& 49.04 \\
\multicolumn{2}{l|}{Google Translate}  &  ~     &   ~    &  ~    &   ~   &   ~     &  16.66 & 59.18 & 46.24& 49.65 \\
\ste{}       &  DeriNet+\ste{}   &  94M   &  138M  &  80k  &  56k  &  35.58  &  15.31 & 69.44 & 44.77& 47.91 \\
Morfessor+\ste{}  &  \ste{}       &  139M  &  86M   &  41k  &  84k  &  26.43  &  14.51 & 68.81 & 43.51& 47.56 \\
\multic{2}{c|}{BPE~shrd~voc} &    95M &  85M   & 56k  &  71k   &  26.78   &  13.79 & 97.94 & 46.44 & 42.49 \\

\end{tabular}

\end{table}

\Tref{tab:dervoc} presents several combinations of linguistically motivated and data-driven segmentation methods.
Since the vocabulary size after Morfessor or DeriNet splitting alone often remains too high, we further split the corpus with BPE or \ste{}.
Unfortunately, none of the setups performs better than the \ste{} baseline.

\subsection{Experiment 2: Allowing Zero Ending}

\Tref{tab:bpet2tbleu} empirically compares \ste{} and variants of BPE. It turns out that \ste{} performs almost 5(!) BLEU
point better than the default BPE. The underscore feature allowing to model zero
suffix almost closes the gap and shared vocabulary also helps a little.


\def\clap#1{\hbox to 0pt{\hss #1\hss}}
\def\yes{\checkmark}
\def\no{-}
\begin{table}[t!]
\caption{BPE vs \ste{} with/without underscore after every (non-final) token of
a sentence and/or shared vocabulary.
Reported scores are avg$\pm$stddev of T2T checkpoints between 275k
and 325k training steps. CharacTER, chrF3 and BEER are multiplied by 100.
}
\label{tab:bpet2tbleu}
\centering
{
\scriptsize
\begin{tabular}{l@{~~}c@{~~}c|r@{~~}r@{~~}r@{~~}r}
\textbf{split} & \textbf{underscore} & \textbf{shared vocab}  & \multicolumn{1}{c}{\textbf{BLEU}}  & \multicolumn{1}{c}{\clap{\textbf{CharacTER}}} & \multicolumn{1}{c}{\textbf{chrF3}} & \multicolumn{1}{c}{\textbf{BEER}} \\
\hline
\ste{}  &  after~every~token       &  \yes{}  &  18.58\plusminus{}0.06  &  61.43\plusminus{}0.68  &  44.80\plusminus{}0.29  &  50.23\plusminus{}0.16  \\  
BPE     &  after~non-final~tokens  &  \yes{}  &  18.24\plusminus{}0.08  &  63.80\plusminus{}0.88  &  44.37\plusminus{}0.24  &  49.84\plusminus{}0.15  \\  
BPE     &  after~non-final~tokens  &  \no{}   &  18.07\plusminus{}0.08  &  63.24\plusminus{}1.98  &  44.21\plusminus{}0.20  &  49.72\plusminus{}0.11  \\  
BPE     &  after~every~token       &  \yes{}  &  13.88\plusminus{}0.18  &  81.84\plusminus{}3.33  &  36.74\plusminus{}0.51  &  42.46\plusminus{}0.51  \\
BPE     &  \no{}                   &  \yes{}  &  13.69\plusminus{}0.66  &  76.72\plusminus{}4.03  &  36.60\plusminus{}0.63  &  42.33\plusminus{}0.60  \\  
BPE     &  \no{}                   &  \no{}   &  13.66\plusminus{}0.38  &  82.66\plusminus{}3.54  &  36.73\plusminus{}0.53  &  42.41\plusminus{}0.56  \\  
\end{tabular}
}
\end{table}

\begin{figure}[t]
\centering
\includegraphics[width=6.5cm]{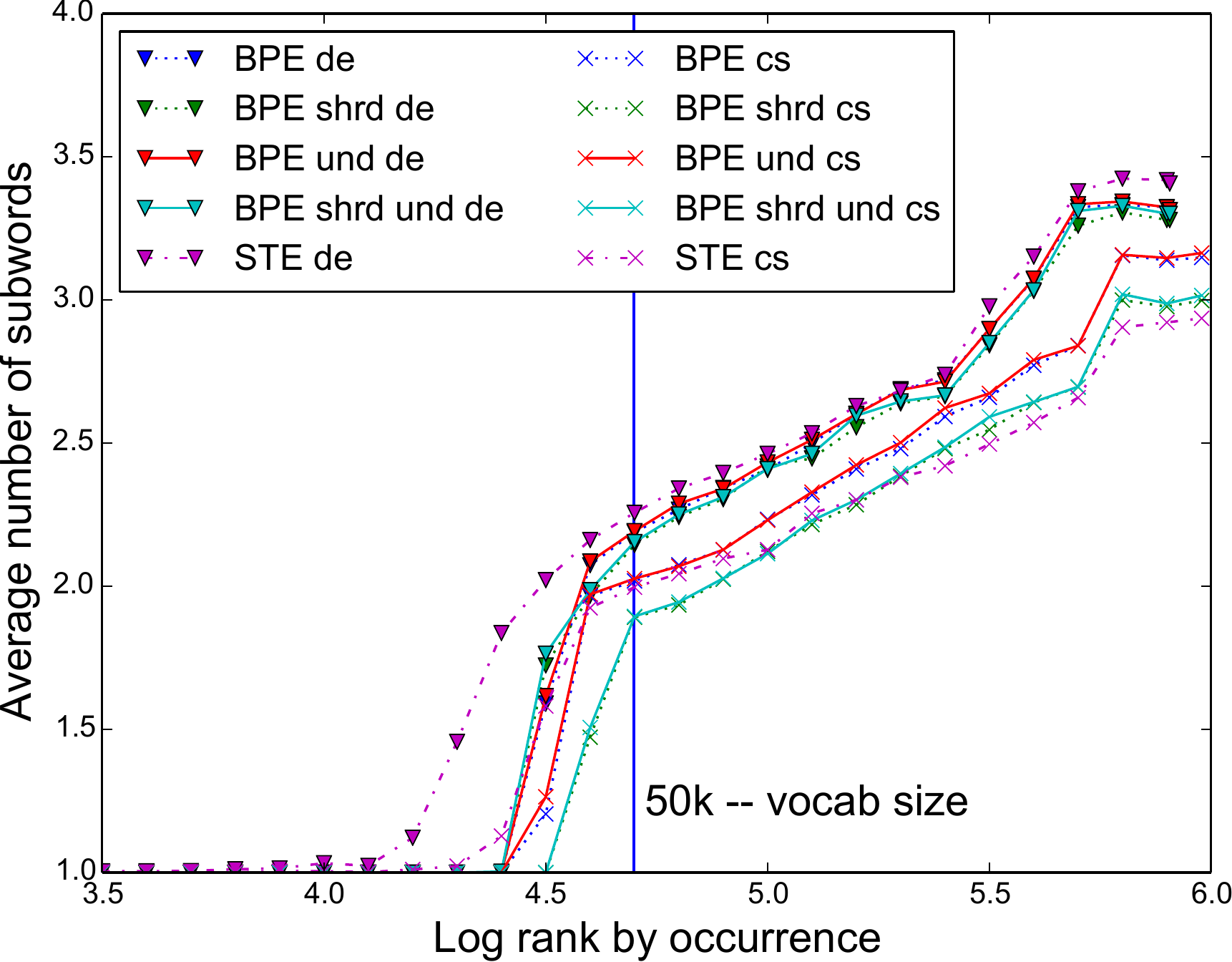}
\caption{
Histogram of number of splits of words based on their frequency rank.
The most common words (left) remain unsplit by all methods, rare words (esp. beyond the 50k vocabulary limit)
are split to more and more subwords.
}
\label{fig:avgboth}
\end{figure}

As \Fref{fig:avgboth} indicates, the difference in performance is not a straightforward consequence of the number of splits generated.
There is basically no difference between BPE with and without
underscore
but shared vocabulary leads to a lower number of splits on the Czech
target side.
We can see that \ste{} in both languages splits words to more parts
than BPE but still performs better. We conclude that the
\ste{} splits allow to exploit morphological behaviour better.


%
%


\section{Discussion}

All our experiments show that our linguistically motivated techniques do not
perform better in machine translation than current state-of-the-art agnostic
methods. Actually, they do not even lead to linguistically adequate splits when
evaluated against a dictionary of word segmentations. This can be caused by the
fact that our new methods are not accurate
enough in splitting words to morphs,
maybe because of the limited size of DeriNet and small amount of training
data for Morfessor, maybe because they don't handle the phonological and orthographical
changes, so the amount of resulting morphs is still very high and most of them are
rare in the data.


One new linguistically adequate feature, the zero suffix mark after all but final
tokens in
the sentence
showed a big
improvement, while adding the mark after every token did not.
This suggests that the Tensor2Tensor NMT model benefits from explicit sentence
ends perhaps more than from a better segmentation, but further investigation is
needed.

%

\section{Conclusion}

We experimented with common linguistically non-informed word segmentation
methods BPE and \stefull{}, and with two linguistically-motivated ones. Neither
Morfessor nor our novel technique relying on DeriNet, a derivational dictionary
for Czech, help. The uninformed methods thus remain the best choice.

Our analysis however shows an important difference in \ste{} and BPE, which
leads to considerably better performance. The same feature (support for zero
suffix) can be utilized in
BPE, giving similar gains.

\XXX{
	Reviewer3:
Further comments:
One would like to know the authors opinion on how the story will
continue: will we need morphology in future and if so, for which
languages/language pairs.
}

%
%
%
\bibliographystyle{splncs04}
\bibliography{biblio}

\end{document}